\documentclass[pmlr]{jmlr} % new name PMLR (Proceedings of Machine Learning Research)

\newcommand{\tagdsmode}{nonproceedings}

\makeatletter
\newcommand{\tagdssubmission}{submission}
\newcommand{\tagdsproceedings}{proceedings}

\ifx\tagdsmode\tagdsproceedings
\else\ifx\tagdsmode\tagdssubmission
  \def\ps@jmlrtps{%
    \let\@mkboth\@gobbletwo
    \def\@oddhead{\scriptsize Under Review at the 2nd Conference on Topology, Algebra, and Geometry in Data Science\hfill}%
    \let\@evenhead\@oddhead
    \def\@oddfoot{}%
    \let\@evenfoot\@oddfoot
  }

\else
  \def\ps@jmlrtps{%
    \let\@mkboth\@gobbletwo
    \def\@oddhead{}%
    \let\@evenhead\@oddhead
    \def\@oddfoot{}%
    \let\@evenfoot\@oddfoot
  }
\fi\fi
\makeatother

\usepackage{longtable}% for long tables

\usepackage{booktabs}
\usepackage[load-configurations=version-1]{siunitx} % newer version

\usepackage{mathtools}
\usepackage{caption}
\newcommand{\RoVE}{\mbox{\textsc{R\kern.1pto\kern-1ptVE}}\xspace}

\theorembodyfont{\upshape}
\theoremheaderfont{\scshape}
\theorempostheader{:}
\theoremsep{\newline}

\jmlrvolume{334}
\jmlryear{2026}
\jmlrworkshop{Topology, Algebra, and Geometry in Data Science}

\title[\RoVE: Rotary Value Embeddings Attention]{\RoVE: Rotary Value Embeddings Attention\\ for Relative Position-dependent Value Pathways}

\ifx\tagdsmode\tagdssubmission

\else
\newcommand{\aff}[1]{$^{#1}$}

\author[García-Castellanos, Weiler \& Bekkers]{
   \Name{Alejandro García-Castellanos\aff{1}} \Email{
a.garciacastellanos@uva.nl} \\
   \Name{Maurice Weiler\aff{2}} \Email{m.weiler.ml@gmail.com}\\
   \Name{Erik J. Bekkers\aff{1}} \Email{e.j.bekkers@uva.nl}\\
   \addr AMLab, University of Amsterdam\aff{1},\,\ \ MIT CSAIL\aff{2}
   }

\fi

\ifx\tagdsmode\tagdsproceedings
\editor{Editor's name}
\fi

\usepackage{caption}

\newcommand{\Rot}[1]{R_{#1}}
\newcommand{\tp}{^{\!\top}}
\newcommand{\Wv}{W_{\!V}}
\newcommand{\Wo}{W_{\!O}}
\newcommand{\Wq}{W_{\!Q}}
\newcommand{\Wk}{W_{\!K}}
\newcommand{\calN}{\mathcal{N}}
\newcommand{\R}{\mathbb{R}}

\newcommand{\softmax}{\operatorname{softmax}}

\newcommand{\kernel}[1]{\psi_{#1}}
\newcommand{\shift}[1]{S^{#1}}

\usepackage[table]{xcolor}
\definecolor{bestcol}{RGB}{180,230,180}
\newcommand{\best}[1]{\cellcolor{bestcol}{#1}}

\begin{document}

\maketitle

\vspace*{-4ex}
\begin{abstract}
Rotary Position Embeddings (\textsc{RoPE}) make attention scores
position-relative but leave the value pathway position-blind:
the message sent by a value token is the same regardless of its distance from the query.
We propose \RoVE, a parameter-free modification that makes values position-sensitive by rotating them simultaneously with keys, and show that it turns \textsc{RoPE} attention into \emph{attentive convolution}.
This new perspective unifies several independent formulations of
the same operation across computer vision, robotics, and modern LLM
architectures. Trained 124M and 354M GPT-2 models show consistent empirical
gains over \textsc{RoPE} on few-shot in-context learning,
out-of-distribution perplexity, and long-context retrieval, with the clearest
improvements on tasks that require long-range aggregation.
\end{abstract}

% =============================================================================
\section{Introduction}
% =============================================================================
Rotary Position Embeddings (\textsc{RoPE})~\citep{su2024roformer} make
attention scores \emph{shift equivariant}:
rotating queries $q_i$ and keys $k_j$ by position dependent rotation matrices $\Rot{i}$ and~$\Rot{j}$ produces the score $q_i\tp\Rot{j-i}k_j/\!\sqrt{d}$, which depends on
positions only through the relative offset $\delta=j-i$. The value pathway, however, is
untouched.
In the language of transformer circuits
(Appendix~\ref{app:circuits} and \citet{elhage2021mathematical}),
% \citep[and Appendix~\ref{app:circuits}]{elhage2021mathematical},
% see in Appendix~\ref{app:circuits},
\textsc{RoPE} biases the QK circuit toward relative positions while leaving the
OV circuit position-blind:
unlike convolution kernels, the channel map $\Wv$ applied to $x_j$ carries no
information about where $x_j$ \mbox{lies relative to the query.}

A natural completion rotates the value at position $j$ by $\Rot{j}$ before
aggregation and inverts by moving the output via $\Rot{i}^{-1}$ to position $i$.
As we show, this replaces the constant value map $\Wv$ with the offset-dependent convolution kernel $\kernel{\delta}=\Rot{\delta}\Wv$ where $\delta=j-i$,
endowing the OV circuit with the same
relative-position sensitivity already present in the QK circuit.

Similar ``\textsc{RoPE}-on-values'' constructions have been independently discovered across several communities.
\cite{miyato2024gta} introduced it
for multi-view novel-view synthesis, encoding geometric relationships between
camera frames, and subsequent work has further extended the same mechanism to
computer vision~\citep{wu2026rayropeprojectiveraypositional,li2025camerasrelativepositionalencoding}
and robotics~\citep{klee2026raven}. In the language-modeling setting,
DeepSeek-V4~\citep{deepseekai2026deepseekv4} arrives at the same operation from a
different direction: its compressed shared-KV architecture causes positional
information to leak from keys into values, and an inverse output rotation needs
to be applied as a corrective measure to maintain the relative position. Each
work motivates the mechanism on application-specific grounds, yet none provides a
structural account of what it does to the attention operator.

We isolate this positional embedding mechanism on the value stream, which we call \RoVE, and provide a theoretical analysis of this modification. We then evaluate it as a standalone module in standard (non-shared-KV)
language models -- a regime in which it has not previously been studied. Our
contributions are:

\begin{itemize}\setlength{\itemsep}{0pt}
  \item \textbf{Structural characterisation:} We show that \RoVE turns
  \textsc{RoPE} attention into an \emph{attentive
  convolution}~\citep{romero2020attentivegroupequivariantconvolutional,NEURIPS2020_15231a7c}:
  the position-blind map $\Wv$ is replaced by the offset-dependent block-Toeplitz kernel
  $\kernel{\delta}=\Rot{\delta}\Wv$, and the matrix mixing operator \citep{hwang2024hydrabidirectionalstatespace}
  acquires gated block-Toeplitz rather than Kronecker structure.

  \item \textbf{Empirical validation:} We train 124M and 354M parameter GPT-2
  language models and evaluate on in-context learning (ICL) benchmarks and
  long-context tasks. \RoVE consistently improves ICL accuracy,
  long-context robustness, and retrieval performance over standard
  \textsc{RoPE} attention.
\end{itemize}

\begin{figure}[t]
\floatconts
  {fig:mixer}
  {\vspace{-15pt}\caption{\textbf{Matrix-mixer view of \textsc{RoPE} and \RoVE.}
  \textsc{RoPE} factorises into
  $(a)$ position-sensitive attention weights and
  % $(b)$ a \emph{constant} value kernel $W_V$,
  % so the channel transformation is shared across all offsets $(c)$.
  $(b)$ a \emph{constant} shared value projection $W_V$ across all offsets $(c)$.
  $(d)$ \RoVE replaces $W_V$ with the offset-indexed block-Toeplitz kernel family $\psi_\delta = R_\delta W_V$,
  $(e)$ producing a gated block-Toeplitz mixer whose kernel diagonals rotate systematically with
  relative offset, the signature of an attentive convolution.\vspace{-10pt}}}
 {\setlength{\jmlrminsubcaptionwidth}{0.15\linewidth}%
    \subfigure[\small{Att.\ weights}]{\label{fig:mixer_attn}%
      \includegraphics[width=0.193\linewidth]{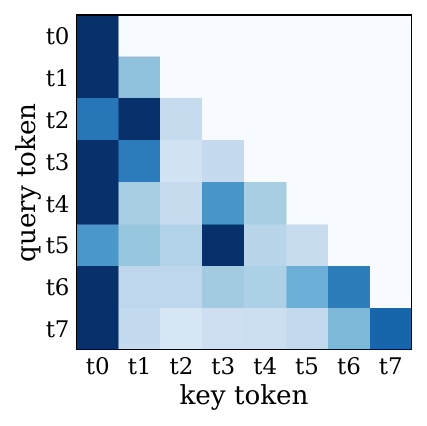}}%
    \hfill%
    \subfigure[\small{\textsc{RoPE} kernel}]{\label{fig:mixer_rope_kernel}%
      \includegraphics[width=0.18\linewidth]{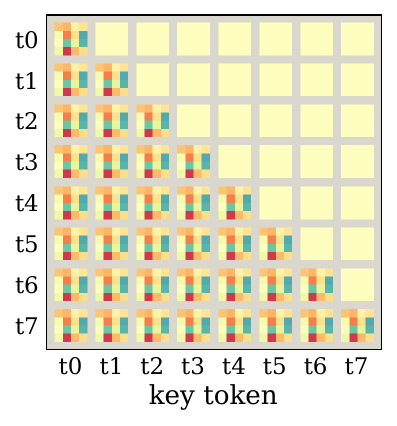}}%
    \hfill%
    \subfigure[\small{\textsc{RoPE} mixer}]{\label{fig:mixer_rope_mixer}%
      \includegraphics[width=0.18\linewidth]{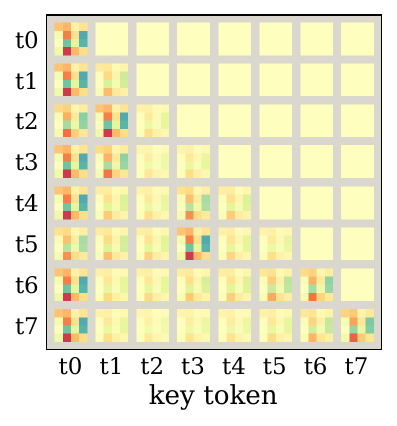}}%
    \hfill%
    \subfigure[\small \RoVE kernel]{\label{fig:mixer_RoVE_kernel}%
      \includegraphics[width=0.18\linewidth]{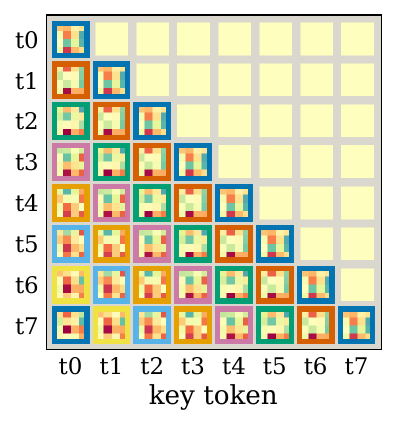}}%
    \hfill%
    \subfigure[\small{\RoVE mixer}]{\label{fig:mixer_RoVE_mixer}%
      \includegraphics[width=0.18\linewidth]{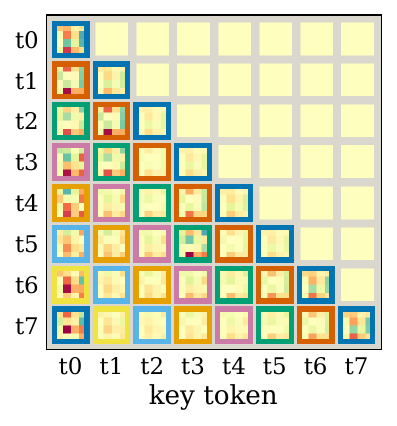}}%
  }
\end{figure}

\vspace{-15pt}
% =============================================================================
\section{Background}
\label{sec:background}
% \vspace{-2pt}
% =============================================================================
\paragraph{Matrix mixers:}
Let $X \in \mathbb{R}^{n \times d}$ be the input feature tensor,
where $n$ is the sequence length and $d$ is the hidden dimension.
We denote by $\operatorname{vec}(X) \in \mathbb{R}^{nd}$ the row-major
(token-major) vectorisation of $X$, stacking the per-token feature vectors as
$\operatorname{vec}(X)=(x_1^\top,\dots,x_n^\top)^\top$.
Any layer of the form $\operatorname{vec}(Y)=\mathcal{M}(X)\operatorname{vec}(X)$,
where $\mathcal{M}(X)$ is an $nd\times nd$ matrix partitioned into $d\times d$ blocks
indexed by token pairs $(i,j)$,
belongs to the \emph{matrix mixer} family~\citep{hwang2024hydrabidirectionalstatespace}.
% The block structure of $\mathcal{M}$ directly encodes inductive biases:
% diagonal blocks mean no token mixing; Toeplitz structure imposes
% shift-equivariance; a Kronecker factorisation decouples the token and channel
% dimensions entirely.

\paragraph{Attentive convolutions:}
A classical $d$-dimensional convolution mixes positions through a fixed
offset-dependent kernel $\kernel{\delta}\in\R^{d\times d}$. \emph{Attentive
convolutions}~\citep{romero2020attentivegroupequivariantconvolutional, NEURIPS2020_15231a7c} replace the fixed weights with
content-dependent scalars  $A(X)_{ij}\in\R$ while keeping the kernel:
\vspace{-5pt}
\begin{align}
  y_i\ =\ \sum\nolimits_{j\in\calN(i)} A(X)_{ij}\;\kernel{j-i}\;x_j,
  \label{eq:attconv}
\end{align}
where $A(X)_{ij}$ gates the contribution of token $j$ to position $i$, and
$\calN(i)$ is a neighbourhood of $i$. The kernel $\kernel{j-i}$ is
\emph{block-Toeplitz}, i.e., constant along each block diagonal, while the mixer is
\emph{gated block-Toeplitz}: its $(i,j)$-th $d\times d$ block of
$\mathcal{M}(X)$ equals $A(X)_{ij}\kernel{j-i}$, so each block diagonal shares
the same kernel, modulated only by the content-dependent scalar gate
$A(X)_{ij}$.

% \paragraph{Standard RoPE attention:}
% \textsc{RoPE}~\citep{su2024roformer} defines block-diagonal orthogonal matrices
% $\Rot{t}\in\mathrm{SO}(d)$ whose $k$-th $2\times 2$ block rotates by angle
% $t\omega_k$ (frequencies $\omega_k=\theta_0^{-2k/d}$).
% Causal \textsc{RoPE} attention computes
% \begin{equation}
%   y_i = \sum_{j\le i} A(X)_{ij}\,\Wv x_j, \qquad
%   A(X)_{ij} = \softmax_{j\le i}\!\left(
%     \tfrac{1}{\sqrt{d}}\,(\Wq x_i)\tp\Rot{j-i}(\Wk x_j)
%   \right),
%   \label{eq:rope-attn}
% \end{equation}
% where $A(X)_{ij}\in\R$ are scalar attention weights. 
% % Because logits depend on
% % absolute positions only through the offset $j-i$, the attention weights are
% % shift-equivariant: shifting the input sequence by $t$, i.e.\
% % $[{t}\rhd X]_i = x_{i-t}$, satisfies $A({t}\rhd X)_{i+t,j+t} = A(X)_{ij}$.
% The mixer factorises as $\mathcal{M}^{\textsc{RoPE}}(X) = A(X)\otimes\Wv$,
% decoupling token routing from channel transformation across independent tensor
% dimensions \citep{elhage2021mathematical}. For further related work see
% Appendix~\ref{app:related}.

\paragraph{Standard \textsc{RoPE} attention:}
Let $\mathcal{N}(i)\subseteq\{1,\dots,n\}$ denote the set of positions visible to query $i$. Typical choices include causal masking ($\calN(i)=\{j\le i\}$), full attention ($\calN(i)\equiv \{1, ..., n\}$), and sparse patterns~\citep{child2019generatinglongsequencessparse}.
\textsc{RoPE}~\citep{su2024roformer} defines block-diagonal rotation matrices $\Rot{t}\in\mathrm{SO}(d)$, where the $m$-th $2\times 2$ block rotates by angle $t\omega_m$ for geometrically spaced frequencies $\omega_m=\theta_0^{-2m/d}$, and computes
\begin{align}
  y_i = \sum\nolimits_{j\in\mathcal{N}(i)} A(X)_{ij}\,\Wv x_j, \qquad
  A(X)_{ij} = \softmax_{j\in\mathcal{N}(i)}\!\left(
    \tfrac{1}{\sqrt{d}}\,(\Wq x_i)\tp\Rot{j-i}(\Wk x_j)
  \right).
  \label{eq:rope-attn}
\end{align}
%
% Because each logit depends on positions only through the offset $\delta=j-i$,
% the attention pattern is shift-equivariant.
% The value pathway, however, is untouched: $\Wv x_j$ carries no information about
% where $j$ lies relative to $i$.
The mixer factorizes as $\mathcal{M}^{\textsc{RoPE}}(X) = A(X)\otimes\Wv$ across tensor
axes, decoupling token routing from channel projections \citep{elhage2021mathematical}.

\paragraph{\textsc{YaRN}:}

% - often training at short context, deploy or fine-tune at longer context lengths
% - out of distribution:  ....
% - stretch frequencies uniformly  ->  destroys local position dependence (high frequencies)
% - YaRN:  ramp function
% - see Appendix X    

% \textsc{RoPE} generalises poorly beyond the training context length, as
% low-frequency components are the first to encounter rotation angles unseen
% during training~\citep{tian2026mrropemixedradixrotaryposition}.
% \textsc{YaRN}~\citep{peng2024yarn} addresses this
% % at inference time
% % via an NTK-by-parts strategy:
% as follows:
% high frequencies are left unchanged,
% low frequencies are fully interpolated,
% and a ramp function blends the two in between,
% enabling effective long-context extension.

\textsc{RoPE}-based models are typically trained at short context lengths but deployed at longer ones, creating an out-of-distribution problem: low-frequency components encounter rotation angles unseen during training~\citep{tian2026mrropemixedradixrotaryposition}. Uniform frequency stretching is a natural fix~\citep{chen2023extendingcontextwindowlarge}, but destroys local position dependence in high-frequency components. \textsc{YaRN}~\citep{peng2024yarn} resolves this via frequency-dependent interpolation, i.e., preserving high frequencies while smoothly scaling the lower ones. This rescaling can be applied post-hoc without any additional training. For further related work see Appendix~\ref{app:related}.

% \vspace{-10pt}

% =============================================================================
\section{Method}
\label{sec:RoVE}
% =============================================================================

\textsc{RoPE} makes attention scores position-relative but leaves the value pathway invariant:
two tokens assigned equal attention weight contribute identically to the output regardless of their offset from the query.
\RoVE extends this by additionally rotating each value into the query's reference frame before aggregation.

\begin{definition}[\RoVE]
\label{def:RoVE}
Let $\mathcal{N}(i)\subseteq\{1,\dots,n\}$ be any neighbourhood function and let $A(X)_{ij}$ be the \textsc{RoPE} attention weights from
\eqref{eq:rope-attn}. \RoVE computes
\begin{equation}
\tilde{y}_i
\,\ =\,\ 
\Rot{i}^{-1}
\sum_{\mathclap{j\in \mathcal{N}(i)}}
A(X)_{ij}\,\Rot{j}\,\Wv x_j
\,\ =\,\ 
\sum_{\mathclap{j\in \mathcal{N}(i)}}
A(X)_{ij}\;
\underbrace{\Rot{j-i}\Wv}_{\scriptstyle\kernel{j-i}}
\,x_j .
\label{eq:RoVE}
\end{equation}
\end{definition}

% \vspace*{-1ex}
\paragraph{Convolution lens:}
Equation~\eqref{eq:RoVE} is an instance of the attentive convolution
\eqref{eq:attconv} with neighbourhood $\mathcal{N}$,
\textsc{RoPE} attention weights, and position-dependent kernel
$\kernel{\delta}=\Rot{\delta}\Wv$ (standard \textsc{RoPE} is recovered by
the degenerate choice $\kernel{\delta}\equiv\Wv$). Crucially, the value
pathway is no longer a single shared map but the tied family
$\{\Rot{\delta}\Wv\}_{\delta}$, so relative position modulates token
\emph{transformation} (OV circuit) as well as token \emph{selection} (QK circuit).

\paragraph{Matrix mixer lens:}
In mixer form, the $(i,j)$-th $d\times d$ block of $\mathcal{M}^{\RoVE}(X)$
equals $A(X)_{ij}\Rot{j-i}\Wv$, replacing the factorised \textsc{RoPE} blocks
$A(X)_{ij}\Wv$.
Consequently, the mixer inherits the \emph{gated block-Toeplitz structure} of attentive
convolutions: the value kernel is block-Toeplitz, modulated by the
content-dependent scalars $A_{ij}$, so blocks along
the same relative-offset diagonal share the same rotated value kernel.
Figure \ref{fig:mixer} visualizes this transition from a constant value kernel to an
offset-indexed family of value kernels. See Appendix~\ref{app:circuits} for further analysis of \RoVE's circuit.

% \paragraph{\rm \textit{Note.}} Further analysis of the method from the local frame, and the transformer circuit lens, as well as efficiency discussion can be found in the Appendix.
\paragraph{Local-frame lens:}
Equation~\eqref{eq:RoVE} admits a clean frame-change interpretation: first, each value $\Wv x_j$
is transformed from its local frame into a shared global frame by $\Rot{j}$, then contributions
are aggregated in the global frame, and finally the result is transformed into the query's
local frame by $\Rot{i}^{-1}$. The effective kernel $\Rot{j-i}\Wv$
is the frame-change operator composed with the learned channel map.
This frame-change perspective is the primary motivation
in~\cite{miyato2024gta},
where features from different views are rotated into a common reference frame
before aggregation. Moreover, in the message-passing view, the transformed values $\Rot{j}\Wv x_j$ can be seen as \emph{tensorial messages}~\citep{lippmann2025canonicalizationtensorialmessagesimprove}.
% Moreover, in the message-passing view, the transformed values $\Rot{j}\Wv x_j$
% are \emph{tensorial messages}~\citep{lippmann2025canonicalizationtensorialmessagesimprove}:
% they carry frame information from source to target,
% so each token communicates both its content and its positional context.

\paragraph{Efficiency:}
\RoVE embeds values analogously to the query/key embeddings in RoPE.
These operations have linear complexity $\mathcal{O}(nd)$ since rotations act independently on $n$ individual tokens and $\frac{d}{2}$ channel pairs --
their computational cost is therefore negligible compared to $\mathcal{O}(nd^2)$ linear and $\mathcal{O}(n^2d)$ attention layers.
Like RoPE, \RoVE is compatible with FlashAttention kernels \citep{dao2022flashattention} since it acts on values and attention outputs before and after the kernel call.
Furthermore, it introduces no additional learned parameters.

\vspace{-10pt}
% =============================================================================
\section{Experiments}
\label{sec:experiments}
% =============================================================================

\paragraph{Setup:}
We evaluate \RoVE as a drop-in replacement for the value pathway in
\textsc{RoPE} attention, training GPT-2-style transformers~\citep{brown2020language}
at small (${\sim}124$M) and medium (${\sim}354$M) scale on
FineWebEdu-10B~\citep{lozhkov2024fineweb-edu} with a 1024-token context.
We evaluate on: \textbf{DCLM-Core}, few-shot ICL accuracy within the training
context~\citep{li2025datacomplmsearchgenerationtraining}; \textbf{OOD perplexity}
at up to $16\times$ context length, with and without
\textsc{YaRN}~\citep{peng2024yarn}; and \textbf{RULER}, long-context
retrieval at 4k/8k tokens scored by NLL~\citep{hsieh2024ruler}.
Full details and more empirical results are shown in Appendix~\ref{app:experiments}.

\begin{table*}[ht]
\centering
\caption{
    Core ICL accuracy and perplexity for the 354M parameter model. Core is measured within
    the 1024-token training context; PPL is measured from 512 to 16384
    tokens. $+\textit{YaRN}$ denotes inference-time interpolation, leaving Core
    unchanged. \colorbox{bestcol}{Green highlight} marks column bests}

\setlength{\tabcolsep}{5pt}
\small
\begin{tabular}{lrrrrrrr}
\toprule
 &  & \multicolumn{6}{c}{PPL ($\downarrow$)} \\
\cmidrule(lr){3-8}
Method & Core ($\uparrow$) & 512 & 1024 & 2048 & 4096 & 8192 & 16384 \\
\midrule
\textsc{RoPE}         & 0.1664        & 17.68        & 15.64        & 293.24        & 840.10        & 1412.76       & 1630.72       \\
$+\textit{YaRN}$       &               & 17.68        & 15.64        & 16.05         & 48.61         & 154.61        & 270.98        \\
\midrule
\RoVE (ours)        & \best{0.1856} & \best{17.52} & \best{15.52} & 133.68        & 311.38        & 458.15        & 583.84        \\
$+\textit{YaRN}$      &               & \best{17.52} & \best{15.52} & \best{15.78}  & \best{18.40}  & \best{35.95}  & \best{124.82} \\
\bottomrule
\end{tabular}
\label{tab:core_ppl_354}
\end{table*}

\paragraph{\RoVE improves in the trained regime:}
Tables~\ref{tab:core_ppl_354} and~\ref{tab:core_ppl_124} show that \RoVE
improves both Core ICL accuracy and in-context perplexity at both scales.
Thus \RoVE provides a useful inductive bias even within the
training distribution, not only at extrapolated lengths.

\paragraph{\RoVE and \textsc{YaRN} are complementary:}
\RoVE substantially reduces OOD perplexity, and \textsc{YaRN} improves both
baselines while leaving their gap largely intact. Relative positioning on the value stream is therefore
not hindered by inference-time frequency interpolation, i.e., the two methods address
complementary aspects of long-context generalization.

% \paragraph{RoVE improves long-context retrieval.}
% Tables~\ref{tab:yarn_ruler_124} and~\ref{tab:yarn_ruler_354} show perplexity
% gains translate to RULER retrieval, with the largest improvements on tasks
% requiring information to be maintained and recombined across the context.
% This aligns with the attentive-convolution view: standard \textsc{RoPE} leaves
% value transformation position-blind, and \RoVE corrects this precisely
% where it matters most.
% \paragraph{RoVE improves long-context retrieval.}
% Tables~\ref{tab:yarn_ruler_124} and~\ref{tab:yarn_ruler_354} show that the
% perplexity gains translate to synthetic long-context retrieval tasks, with
% the strongest improvements on tasks that require maintaining and recombining
% information across the context. These results align with the attentive
% convolution view. Standard \textsc{RoPE} makes attention weights
% position-aware but leaves value transformation position-blind; \RoVE
% corrects this by aligning selected information according to relative position
% before aggregation. Retrieval gains are largest precisely where this matters
% most.

\begin{table*}[ht]
\centering
\caption{
    RULER long-context retrieval (NLL; lower is better) for the 354M parameter model.
    Tasks: Common Word Extraction (CWE), multi-key Needle-in-a-Haystack (NIAH),
    Question Answering (QA), Variable Tracking (VT). Avg is the unweighted mean
    over all eight task/length cells. $+\textit{YaRN}$ denotes inference-time interpolation. \colorbox{bestcol}{Green highlight} marks column bests.
}
\small
\setlength{\tabcolsep}{4pt}
\resizebox{\linewidth}{!}{%
\begin{tabular}{lccccccccc}
\toprule
 & \multicolumn{2}{c}{CWE ($\downarrow$)} & \multicolumn{2}{c}{NIAH ($\downarrow$)} & \multicolumn{2}{c}{QA ($\downarrow$)} & \multicolumn{2}{c}{VT ($\downarrow$)} &  \\
\cmidrule(lr){2-3}\cmidrule(lr){4-5}\cmidrule(lr){6-7}\cmidrule(lr){8-9}
Method & 4k & 8k & 4k & 8k & 4k & 8k & 4k & 8k & Avg ($\downarrow$)\\
\midrule
\textsc{RoPE}    & $8.87{\scriptstyle\pm0.46}$        & $9.15{\scriptstyle\pm0.44}$        & $10.90{\scriptstyle\pm1.25}$       & $10.40{\scriptstyle\pm1.22}$       & $8.83{\scriptstyle\pm1.84}$        & $9.18{\scriptstyle\pm1.64}$        & $10.12{\scriptstyle\pm0.53}$       & $10.22{\scriptstyle\pm0.38}$       & $9.71{\scriptstyle\pm0.97}$        \\
$+\textit{YaRN}$ & \best{$4.31{\scriptstyle\pm0.47}$} & $5.71{\scriptstyle\pm0.41}$        & $7.61{\scriptstyle\pm1.47}$        & $9.46{\scriptstyle\pm1.27}$        & $6.40{\scriptstyle\pm2.73}$        & $7.43{\scriptstyle\pm2.14}$        & $4.53{\scriptstyle\pm0.51}$        & $7.50{\scriptstyle\pm0.51}$        & $6.62{\scriptstyle\pm1.19}$        \\
\midrule
\RoVE (ours)    & $7.17{\scriptstyle\pm0.37}$        & $7.93{\scriptstyle\pm0.39}$        & $9.51{\scriptstyle\pm1.48}$        & $10.14{\scriptstyle\pm1.43}$       & $8.25{\scriptstyle\pm2.14}$        & $8.71{\scriptstyle\pm2.03}$        & $7.93{\scriptstyle\pm0.51}$        & $8.31{\scriptstyle\pm0.39}$        & $8.50{\scriptstyle\pm1.09}$        \\
$+\textit{YaRN}$ & $4.39{\scriptstyle\pm0.60}$        & \best{$4.62{\scriptstyle\pm0.54}$} & \best{$3.63{\scriptstyle\pm1.16}$} & \best{$5.16{\scriptstyle\pm1.21}$} & \best{$5.61{\scriptstyle\pm3.23}$} & \best{$6.05{\scriptstyle\pm2.76}$} & \best{$2.11{\scriptstyle\pm0.32}$} & \best{$3.10{\scriptstyle\pm0.41}$} & \best{$4.33{\scriptstyle\pm1.28}$} \\
\bottomrule
\end{tabular}}
\label{tab:yarn_ruler_354}
\end{table*}

\paragraph{\RoVE improves long-context retrieval:}
Tables~\ref{tab:yarn_ruler_354} and~\ref{tab:yarn_ruler_124} show that perplexity gains translate to synthetic long-context retrieval tasks, with the strongest improvements on tasks requiring information to be maintained and recombined across the context. These results align with the attentive
convolution view. Standard \textsc{RoPE} makes attention weights
position-aware but leaves value transformation position-blind; \RoVE
closes this gap by aligning selected information according to relative position
before aggregation. 

\vspace{-7pt}

\section{Conclusion}
\label{sec:conclusion}

% \RoVE turns \textsc{RoPE} attention into an attentive convolution by replacing
% the position-blind map $\Wv$ with the offset-indexed kernel
% $\kernel{\delta}=\Rot{\delta}\Wv$, converting the attention mixer from
% Kronecker to block-Toeplitz structure and endowing the OV circuit with the
% same relative-position sensitivity already present in the QK circuit.

We propose \RoVE, an extension of \textsc{RoPE} attention that turns it
into an attentive convolution by replacing the position-blind map $\Wv$ with
the offset-indexed block-Toeplitz kernel $\kernel{\delta}=\Rot{\delta}\Wv$, converting the
attention mixer from Kronecker to gated block-Toeplitz structure and endowing the OV
circuit with the same relative-position sensitivity already present in the QK circuit.

This parameter-free modification consistently improves upon \textsc{RoPE}
across model scales, in-context evaluation, out-of-distribution perplexity, and
RULER retrieval.
Gains are largest with \textsc{YaRN}, confirming the two methods are complementary.
Since \RoVE leaves attention logits unchanged and is compatible with
efficient attention kernels, relative-position-aware values stand as a robust
structural bias for LLMs. For further discussion see Appendix~\ref{sec:discussion}.

% We view \RoVE as a minimal instance of a broader class of mechanisms where attention controls not only which tokens are selected, but how their representations are aligned before aggregation. 

\acks{
Alejandro García Castellanos is funded by the Hybrid Intelligence Center, a 10-year programme funded through the research programme Gravitation which is (partly) financed by the Dutch Research Council (NWO). This publication is part of the project SIGN with file number VI.Vidi.233.220 of the research programme Vidi which is (partly) financed by the Dutch Research Council (NWO) under the grant \url{https://doi.org/10.61686/PKQGZ71565}.
}

% \newpage

\bibliography{references}

\begin{thebibliography}{39}
\providecommand{\natexlab}[1]{#1}
\providecommand{\url}[1]{\texttt{#1}}
\expandafter\ifx\csname urlstyle\endcsname\relax
  \providecommand{\doi}[1]{doi: #1}\else
  \providecommand{\doi}{doi: \begingroup \urlstyle{rm}\Url}\fi

\bibitem[Arora et~al.(2024)Arora, Eyuboglu, Timalsina, Johnson, Poli, Zou, Rudra, and R{\'e}]{arora2024zoology}
Simran Arora, Sabri Eyuboglu, Aman Timalsina, Isys Johnson, Michael Poli, James~Y Zou, Atri Rudra, and Christopher R{\'e}.
\newblock Zoology: Measuring and improving recall in efficient language models.
\newblock In \emph{International conference on learning representations}, volume 2024, pages 15664--15730, 2024.

\bibitem[Barbero et~al.(2024)Barbero, Vitvitskyi, Perivolaropoulos, Pascanu, and Veli{\v{c}}kovi{\'c}]{barbero2024round}
Federico Barbero, Alex Vitvitskyi, Christos Perivolaropoulos, Razvan Pascanu, and Petar Veli{\v{c}}kovi{\'c}.
\newblock Round and round we go! what makes rotary positional encodings useful?
\newblock \emph{arXiv preprint arXiv:2410.06205}, 2024.

\bibitem[bloc97(2023)]{ntk}
bloc97.
\newblock Ntk-aware scaled rope allows llama models to have longer context windows.
\newblock \url{https://www.reddit.com/r/LocalLLaMA/comments/14lz7j5/ntkaware_scaled_rope_allows_llama_models_to_have/}, 2023.

\bibitem[Brown et~al.(2020)Brown, Mann, Ryder, Subbiah, Kaplan, Dhariwal, Neelakantan, Shyam, Sastry, Askell, Agarwal, Herbert-Voss, Krueger, Henighan, Child, Ramesh, Ziegler, Wu, Winter, Hesse, Chen, Sigler, Litwin, Gray, Chess, Clark, Berner, McCandlish, Radford, Sutskever, and Amodei]{brown2020language}
Tom~B. Brown, Benjamin Mann, Nick Ryder, Melanie Subbiah, Jared Kaplan, Prafulla Dhariwal, Arvind Neelakantan, Pranav Shyam, Girish Sastry, Amanda Askell, Sandhini Agarwal, Ariel Herbert-Voss, Gretchen Krueger, Tom Henighan, Rewon Child, Aditya Ramesh, Daniel~M. Ziegler, Jeffrey Wu, Clemens Winter, Christopher Hesse, Mark Chen, Eric Sigler, Mateusz Litwin, Scott Gray, Benjamin Chess, Jack Clark, Christopher Berner, Sam McCandlish, Alec Radford, Ilya Sutskever, and Dario Amodei.
\newblock Language models are few-shot learners.
\newblock \emph{Advances in neural information processing systems}, 33:\penalty0 1877--1901, 2020.

\bibitem[Chen et~al.(2023)Chen, Wong, Chen, and Tian]{chen2023extendingcontextwindowlarge}
Shouyuan Chen, Sherman Wong, Liangjian Chen, and Yuandong Tian.
\newblock Extending context window of large language models via positional interpolation.
\newblock \emph{arXiv preprint arXiv:2306.15595}, 2023.

\bibitem[Chen et~al.(2025)Chen, Lv, Luan, Wang, and Liu]{chen2025hope}
Yuhan Chen, Ang Lv, Jian Luan, Bin Wang, and Wei Liu.
\newblock Hope: A novel positional encoding without long-term decay for enhanced context awareness and extrapolation.
\newblock In \emph{Proceedings of the 63rd Annual Meeting of the Association for Computational Linguistics (Volume 1: Long Papers)}, pages 23044--23056, 2025.

\bibitem[Chi et~al.(2022)Chi, Fan, Ramadge, and Rudnicky]{chi2022kerple}
Ta-Chung Chi, Ting-Han Fan, Peter~J Ramadge, and Alexander Rudnicky.
\newblock Kerple: Kernelized relative positional embedding for length extrapolation.
\newblock \emph{Advances in Neural Information Processing Systems}, 35:\penalty0 8386--8399, 2022.

\bibitem[Child et~al.(2019)Child, Gray, Radford, and Sutskever]{child2019generatinglongsequencessparse}
Rewon Child, Scott Gray, Alec Radford, and Ilya Sutskever.
\newblock Generating long sequences with sparse transformers.
\newblock \emph{arXiv preprint arXiv:1904.10509}, 2019.

\bibitem[Cordonnier et~al.(2019)Cordonnier, Loukas, and Jaggi]{cordonnier2019relationship}
Jean-Baptiste Cordonnier, Andreas Loukas, and Martin Jaggi.
\newblock On the relationship between self-attention and convolutional layers.
\newblock \emph{arXiv preprint arXiv:1911.03584}, 2019.

\bibitem[Dai et~al.(2019)Dai, Yang, Yang, Carbonell, Le, and Salakhutdinov]{dai2019transformer}
Zihang Dai, Zhilin Yang, Yiming Yang, Jaime~G Carbonell, Quoc Le, and Ruslan Salakhutdinov.
\newblock Transformer-xl: Attentive language models beyond a fixed-length context.
\newblock In \emph{Proceedings of the 57th annual meeting of the association for computational linguistics}, pages 2978--2988, 2019.

\bibitem[Dao et~al.(2022)Dao, Fu, Ermon, Rudra, and R{\'e}]{dao2022flashattention}
Tri Dao, Dan Fu, Stefano Ermon, Atri Rudra, and Christopher R{\'e}.
\newblock Flashattention: Fast and memory-efficient exact attention with io-awareness.
\newblock \emph{Advances in neural information processing systems}, 35:\penalty0 16344--16359, 2022.

\bibitem[DeepSeek-AI(2026)]{deepseekai2026deepseekv4}
DeepSeek-AI.
\newblock Deepseek-v4: Towards highly efficient million-token context intelligence, 2026.

\bibitem[Ding et~al.(2024)Ding, Zhang, Zhang, Xu, Shang, Xu, Yang, and Yang]{ding2024longrope}
Yiran Ding, Li~Lyna Zhang, Chengruidong Zhang, Yuanyuan Xu, Ning Shang, Jiahang Xu, Fan Yang, and Mao Yang.
\newblock Longrope: Extending llm context window beyond 2 million tokens.
\newblock \emph{arXiv preprint arXiv:2402.13753}, 2024.

\bibitem[Elhage et~al.(2021)Elhage, Nanda, Olsson, Henighan, Joseph, Mann, Askell, Bai, Chen, Conerly, DasSarma, Drain, Ganguli, Hatfield-Dodds, Hernandez, Jones, Kernion, Lovitt, Ndousse, Amodei, Brown, Clark, Kaplan, McCandlish, and Olah]{elhage2021mathematical}
Nelson Elhage, Neel Nanda, Catherine Olsson, Tom Henighan, Nicholas Joseph, Ben Mann, Amanda Askell, Yuntao Bai, Anna Chen, Tom Conerly, Nova DasSarma, Dawn Drain, Deep Ganguli, Zac Hatfield-Dodds, Danny Hernandez, Andy Jones, Jackson Kernion, Liane Lovitt, Kamal Ndousse, Dario Amodei, Tom Brown, Jack Clark, Jared Kaplan, Sam McCandlish, and Chris Olah.
\newblock A mathematical framework for transformer circuits.
\newblock \emph{Transformer Circuits Thread}, 2021.
\newblock https://transformer-circuits.pub/2021/framework/index.html.

\bibitem[Fu et~al.(2023{\natexlab{a}})Fu, Dao, Saab, Thomas, Rudra, and R{\'e}]{dao2022hungry}
Daniel~Y. Fu, Tri Dao, Khaled~K. Saab, Armin~W. Thomas, Atri Rudra, and Christopher R{\'e}.
\newblock Hungry {H}ungry {H}ippos: Towards language modeling with state space models.
\newblock In \emph{International Conference on Learning Representations}, 2023{\natexlab{a}}.

\bibitem[Fu et~al.(2023{\natexlab{b}})Fu, Epstein, Nguyen, Thomas, Zhang, Dao, Rudra, and R{\'e}]{fu2023simple}
Daniel~Y. Fu, Elliot~L. Epstein, Eric Nguyen, Armin~W. Thomas, Michael Zhang, Tri Dao, Atri Rudra, and Christopher R{\'e}.
\newblock Simple hardware-efficient long convolutions for sequence modeling.
\newblock \emph{arXiv preprint arXiv:2302.06646}, 2023{\natexlab{b}}.

\bibitem[Fuchs et~al.(2020)Fuchs, Worrall, Fischer, and Welling]{NEURIPS2020_15231a7c}
Fabian Fuchs, Daniel Worrall, Volker Fischer, and Max Welling.
\newblock {SE}(3)-{Transformers}: {3D} {Roto}-{Translation} {Equivariant} {Attention} {Networks}.
\newblock In \emph{Advances in {Neural} {Information} {Processing} {Systems}}, volume~33, pages 1970--1981. Curran Associates, Inc., 2020.
\newblock URL \url{https://proceedings.neurips.cc/paper_files/paper/2020/hash/15231a7ce4ba789d13b722cc5c955834-Abstract.html}.

\bibitem[Gopalakrishnan et~al.(2025)Gopalakrishnan, Csord{\'a}s, Schmidhuber, and Mozer]{gopalakrishnan2025decoupling}
Anand Gopalakrishnan, Robert Csord{\'a}s, J{\"u}rgen Schmidhuber, and Michael~C Mozer.
\newblock Decoupling the" what" and" where" with polar coordinate positional embeddings.
\newblock \emph{arXiv preprint arXiv:2509.10534}, 2025.

\bibitem[Hsieh et~al.(2024)Hsieh, Sun, Kriman, Acharya, Rekesh, Jia, Zhang, and Ginsburg]{hsieh2024ruler}
Cheng-Ping Hsieh, Simeng Sun, Samuel Kriman, Shantanu Acharya, Dima Rekesh, Fei Jia, Yang Zhang, and Boris Ginsburg.
\newblock Ruler: What's the real context size of your long-context language models?
\newblock \emph{arXiv preprint arXiv:2404.06654}, 2024.

\bibitem[Hwang et~al.(2024)Hwang, Lahoti, Dao, and Gu]{hwang2024hydrabidirectionalstatespace}
Sukjun Hwang, Aakash Lahoti, Tri Dao, and Albert Gu.
\newblock Hydra: {Bidirectional} {State} {Space} {Models} {Through} {Generalized} {Matrix} {Mixers}, July 2024.
\newblock URL \url{http://arxiv.org/abs/2407.09941}.
\newblock arXiv:2407.09941 [cs].

\bibitem[Klee et~al.(2026)Klee, Hu, Cole, Tian, Wang, Platt, and Walters]{klee2026raven}
David Klee, Boce Hu, Andrew Cole, Heng Tian, Dian Wang, Robert Platt, and Robin Walters.
\newblock {RAVEN}: End-to-end equivariant robot learning with {RGB} cameras.
\newblock In \emph{The Fourteenth International Conference on Learning Representations}, 2026.
\newblock URL \url{https://openreview.net/forum?id=z8BN7KyaPl}.

\bibitem[Li et~al.(2024{\natexlab{a}})Li, Fang, Smyrnis, Ivgi, Jordan, Gadre, Bansal, Guha, Keh, Arora, Garg, Xin, Muennighoff, Heckel, Mercat, Chen, Gururangan, Wortsman, Albalak, Bitton, Nezhurina, Abbas, Hsieh, Ghosh, Gardner, Kilian, Zhang, Shao, Pratt, Sanyal, Ilharco, Daras, Marathe, Gokaslan, Zhang, Chandu, Nguyen, Vasiljevic, Kakade, Song, Sanghavi, Faghri, Oh, Zettlemoyer, Lo, El-Nouby, Pouransari, Toshev, Wang, Groeneveld, Soldaini, Koh, Jitsev, Kollar, Dimakis, Carmon, Dave, Schmidt, and Shankar]{li2025datacomplmsearchgenerationtraining}
Jeffrey Li, Alex Fang, Georgios Smyrnis, Maor Ivgi, Matt Jordan, Samir Gadre, Hritik Bansal, Etash Guha, Sedrick Keh, Kushal Arora, Saurabh Garg, Rui Xin, Niklas Muennighoff, Reinhard Heckel, Jean Mercat, Mayee Chen, Suchin Gururangan, Mitchell Wortsman, Alon Albalak, Yonatan Bitton, Marianna Nezhurina, Amro Abbas, Cheng-Yu Hsieh, Dhruba Ghosh, Josh Gardner, Maciej Kilian, Hanlin Zhang, Rulin Shao, Sarah Pratt, Sunny Sanyal, Gabriel Ilharco, Giannis Daras, Kalyani Marathe, Aaron Gokaslan, Jieyu Zhang, Khyathi Chandu, Thao Nguyen, Igor Vasiljevic, Sham Kakade, Shuran Song, Sujay Sanghavi, Fartash Faghri, Sewoong Oh, Luke Zettlemoyer, Kyle Lo, Alaaeldin El-Nouby, Hadi Pouransari, Alexander Toshev, Stephanie Wang, Dirk Groeneveld, Luca Soldaini, Pang~Wei Koh, Jenia Jitsev, Thomas Kollar, Alexandros~G. Dimakis, Yair Carmon, Achal Dave, Ludwig Schmidt, and Vaishaal Shankar.
\newblock Datacomp-lm: In search of the next generation of training sets for language models.
\newblock \emph{Advances in Neural Information Processing Systems}, 37:\penalty0 14200--14282, 2024{\natexlab{a}}.

\bibitem[Li et~al.(2026)Li, Yi, Liu, Gao, Ma, and Kanazawa]{li2025camerasrelativepositionalencoding}
Ruilong Li, Brent Yi, Junchen Liu, Hang Gao, Yi~Ma, and Angjoo Kanazawa.
\newblock Cameras as relative positional encoding.
\newblock \emph{Advances in Neural Information Processing Systems}, 38:\penalty0 15984--16009, 2026.

\bibitem[Li et~al.(2024{\natexlab{b}})Li, You, Guruganesh, Ainslie, Ontanon, Zaheer, Sanghai, Yang, Kumar, and Bhojanapalli]{li2024functional}
Shanda Li, Chong You, Guru Guruganesh, Joshua Ainslie, Santiago Ontanon, Manzil Zaheer, Sumit Sanghai, Yiming Yang, Sanjiv Kumar, and Srinadh Bhojanapalli.
\newblock Functional interpolation for relative positions improves long context transformers.
\newblock In \emph{International Conference on Learning Representations}, volume 2024, pages 11303--11328, 2024{\natexlab{b}}.

\bibitem[Lippmann et~al.(2025)Lippmann, Gerhartz, Remme, and Hamprecht]{lippmann2025canonicalizationtensorialmessagesimprove}
Peter Lippmann, Gerrit Gerhartz, Roman Remme, and Fred~A Hamprecht.
\newblock Beyond canonicalization: How tensorial messages improve equivariant message passing.
\newblock In \emph{International Conference on Learning Representations}, volume 2025, pages 88067--88087, 2025.

\bibitem[Lozhkov et~al.(2024)Lozhkov, Ben~Allal, von Werra, and Wolf]{lozhkov2024fineweb-edu}
Anton Lozhkov, Loubna Ben~Allal, Leandro von Werra, and Thomas Wolf.
\newblock Fineweb-edu: the finest collection of educational content, 2024.
\newblock URL \url{https://huggingface.co/datasets/HuggingFaceFW/fineweb-edu}.

\bibitem[Miyato et~al.(2024)Miyato, Jaeger, Welling, and Geiger]{miyato2024gta}
Takeru Miyato, Bernhard Jaeger, Max Welling, and Andreas Geiger.
\newblock Gta: A geometry-aware attention mechanism for multi-view transformers.
\newblock In \emph{International Conference on Learning Representations}, volume 2024, pages 8172--8208, 2024.

\bibitem[nanoGPT(2022)]{karpathy2022nanogpt}
nanoGPT.
\newblock nanogpt.
\newblock \url{https://github.com/karpathy/nanoGPT}, 2022.
\newblock GitHub repository.

\bibitem[Peng et~al.(2023)Peng, Alcaide, Anthony, Albalak, Arcadinho, Cao, Cheng, Chung, Grella, Kiran~GV, He, Hou, Kazienko, Kocon, and Kong]{peng2023rwkv}
Bo~Peng, Eric Alcaide, Quentin Anthony, Alon Albalak, Samuel Arcadinho, Huanqi Cao, Xin Cheng, Michael Chung, Matteo Grella, Kranthi Kiran~GV, Xuzheng He, Haowen Hou, Przemyslaw Kazienko, Jan Kocon, and Jiaming et~al. Kong.
\newblock Rwkv: Reinventing rnns for the transformer era.
\newblock \emph{arXiv:2305.13048}, 2023.

\bibitem[Peng et~al.(2024)Peng, Quesnelle, Fan, and Shippole]{peng2024yarn}
Bowen Peng, Jeffrey Quesnelle, Honglu Fan, and Enrico Shippole.
\newblock Yarn: Efficient context window extension of large language models.
\newblock In \emph{International Conference on Learning Representations}, volume 2024, pages 31932--31951, 2024.

\bibitem[Poli et~al.(2023)Poli, Massaroli, Nguyen, Fu, Dao, Baccus, Bengio, Ermon, and Ré]{poli2023hyenahierarchylargerconvolutional}
Michael Poli, Stefano Massaroli, Eric Nguyen, Daniel~Y. Fu, Tri Dao, Stephen Baccus, Yoshua Bengio, Stefano Ermon, and Christopher Ré.
\newblock Hyena {Hierarchy}: {Towards} {Larger} {Convolutional} {Language} {Models}, April 2023.
\newblock URL \url{http://arxiv.org/abs/2302.10866}.
\newblock arXiv:2302.10866 [cs].

\bibitem[Press et~al.(2021)Press, Smith, and Lewis]{press2022alibi}
Ofir Press, Noah~A Smith, and Mike Lewis.
\newblock Train short, test long: Attention with linear biases enables input length extrapolation.
\newblock \emph{arXiv preprint arXiv:2108.12409}, 2021.

\bibitem[Raffel et~al.(2020)Raffel, Shazeer, Roberts, Lee, Narang, Matena, Zhou, Li, and Liu]{raffel2020t5}
Colin Raffel, Noam Shazeer, Adam Roberts, Katherine Lee, Sharan Narang, Michael Matena, Yanqi Zhou, Wei Li, and Peter~J Liu.
\newblock Exploring the limits of transfer learning with a unified text-to-text transformer.
\newblock \emph{Journal of machine learning research}, 21\penalty0 (140):\penalty0 1--67, 2020.

\bibitem[Romero et~al.(2020)Romero, Bekkers, Tomczak, and Hoogendoorn]{romero2020attentivegroupequivariantconvolutional}
David~W. Romero, Erik~J. Bekkers, Jakub~M. Tomczak, and Mark Hoogendoorn.
\newblock Attentive {Group} {Equivariant} {Convolutional} {Networks}, June 2020.
\newblock URL \url{http://arxiv.org/abs/2002.03830}.
\newblock arXiv:2002.03830 [cs].

\bibitem[Su et~al.(2024)Su, Ahmed, Lu, Pan, Bo, and Liu]{su2024roformer}
Jianlin Su, Murtadha Ahmed, Yu~Lu, Shengfeng Pan, Wen Bo, and Yunfeng Liu.
\newblock Roformer: Enhanced transformer with rotary position embedding.
\newblock \emph{Neurocomputing}, 568:\penalty0 127063, 2024.

\bibitem[Tian et~al.(2026)Tian, Zhu, Liu, Wang, and Wang]{tian2026mrropemixedradixrotaryposition}
Qingyuan Tian, Wenhong Zhu, Xiaoran Liu, Xiaofeng Wang, and Rui Wang.
\newblock Mrrope: Mixed-radix rotary position embedding.
\newblock \emph{arXiv preprint arXiv:2601.22181}, 2026.

\bibitem[Vaswani et~al.(2017)Vaswani, Shazeer, Parmar, Uszkoreit, Jones, Gomez, Kaiser, and Polosukhin]{vaswani2017attention}
Ashish Vaswani, Noam Shazeer, Niki Parmar, Jakob Uszkoreit, Llion Jones, Aidan~N Gomez, {\L}ukasz Kaiser, and Illia Polosukhin.
\newblock Attention is all you need.
\newblock \emph{Advances in neural information processing systems}, 30, 2017.

\bibitem[Wu et~al.(2026)Wu, Jeon, Chang, Tuzel, and Tulsiani]{wu2026rayropeprojectiveraypositional}
Yu~Wu, Minsik Jeon, Jen-Hao~Rick Chang, Oncel Tuzel, and Shubham Tulsiani.
\newblock {RayRoPE}: {Projective} {Ray} {Positional} {Encoding} for {Multi}-view {Attention}, January 2026.
\newblock URL \url{http://arxiv.org/abs/2601.15275}.
\newblock arXiv:2601.15275 [cs].

\bibitem[Zheng et~al.(2025)Zheng, Gao, Shi, Xiong, Sun, Li, Huang, Ren, Ng, Jiang, et~al.]{zheng2025dape}
Chuanyang Zheng, Yihang Gao, Han Shi, Jing Xiong, Jiankai Sun, Jingyao Li, Minbin Huang, Xiaozhe Ren, Michael Ng, Xin Jiang, et~al.
\newblock Dape v2: Process attention score as feature map for length extrapolation.
\newblock In \emph{Proceedings of the 63rd Annual Meeting of the Association for Computational Linguistics (Volume 1: Long Papers)}, pages 10628--10666, 2025.

\end{thebibliography}

\newpage

\appendix
% =============================================================================
\section{Full Experimental Results}
\label{app:experiments}
% =============================================================================

The code, including scripts for generating the results, is available in the following public GitHub repository: \url{https://github.com/AGarciaCast/RoVE}.

\subsection{Setup}
\paragraph{Models:}
We train two GPT-2-style transformers in the nanoGPT
framework~\citep{karpathy2022nanogpt}. The \emph{small} model
(${\approx}124$M parameters) has 12 layers, 12 attention heads, and
embedding dimension 768; the \emph{medium} model (${\approx}354$M
parameters) has 24 layers, 16 heads, and embedding dimension 1024.
Both models share a GPT-2 BPE vocabulary of 50\,304 tokens,
pre-layer normalisation, GELU activations, and base RoPE frequency
$\theta_0 = 10\,000$. The \emph{only} architectural difference between
the \textsc{RoPE} and \RoVE conditions is the value pathway;
all other architectural and training hyperparameters are held fixed.

\paragraph{Training:}
Both models are trained for one epoch on FineWebEdu-10B
(${\approx}10$B tokens of educational web text tokenised with the GPT-2
tiktoken encoder)~\citep{lozhkov2024fineweb-edu}, with a sequence length
of 1024 tokens. We use a total batch size of $2^{19} = 524\,288$ tokens,
accumulated via gradient accumulation over micro-batches of 32 sequences
per GPU (small) and 16 (medium), distributed across four NVIDIA H100 GPUs
with PyTorch DDP. We optimise with AdamW ($\beta = (0.9, 0.95)$, weight
decay $0.1$, gradient clipping at norm $1.0$) in \texttt{bfloat16}. The
learning rate follows a cosine decay from $6 \times 10^{-4}$ to
$6 \times 10^{-5}$ after a 715-step linear warm-up, over 19\,073 gradient
steps in total.

\paragraph{Evaluation:}
We evaluate both models on three benchmarks.

\begin{itemize}

\item \textbf{Core ICL accuracy.} We report in-context learning accuracy
on the DCLM-Core
benchmark~\citep{li2025datacomplmsearchgenerationtraining}, a
diverse suite of few-shot tasks spanning multiple-choice, Winograd
schema, and language-modelling formats. Per-task accuracy is centred on
the random baseline and normalised to $[0,1]$, and the Core score is the
mean across tasks. All evaluation uses the GPT-2 tiktoken tokeniser,
matching training.

\item \textbf{Out-of-distribution perplexity.} We evaluate on the
FineWebEdu-10B held-out validation split at context lengths $L \in
\{512, 1024, 2048, 4096, 8192, 16384\}$ using a sliding window with
stride 512 tokens. Only the final $\min(L, 512)$ tokens per window are
scored, so each reported perplexity reflects next-token prediction
conditioned on $L$ tokens of preceding context. We additionally apply \textsc{YaRN}~\citep{peng2024yarn} positional interpolation
at inference time without any fine-tuning, where the frequency modulation
is applied to all rotation matrices, covering both the QK- and
OV-circuits.

\item \textbf{RULER long-context retrieval.} We evaluate on four RULER
synthetic tasks~\citep{hsieh2024ruler}, namely Common Word Extraction (CWE),
multi-key Needle-in-a-Haystack (NIAH), Question Answering (QA), and
Variable Tracking (VT), at context lengths 4\,096 and 8\,192 tokens
with 500 samples each, constructed with the NVIDIA RULER pipeline using
the GPT-2 tokeniser. Because our models are base language models without
instruction tuning, candidate answers are ranked by negative
log-likelihood (NLL, where lower values indicate higher probability assigned to
the correct answer). We report mean $\pm$ standard deviation over the 500
samples.

\end{itemize}

\subsection{Results}
\label{app:results}

\paragraph{In-distribution performance:}
Tables~\ref{tab:core_ppl_124} and~\ref{tab:core_ppl_354} report Core ICL
accuracy and perplexity within the training context length ($\le 1024$
tokens) for both scales. \RoVE improves Core from $0.1375$ to
$0.1416$ at 124M and from $0.1664$ to $0.1856$ at 354M. Correspondingly,
perplexity at 512 and 1024 tokens decreases from $25.23$/$22.37$ to
$25.05$/$22.30$ (124M) and from $17.68$/$15.64$ to $17.52$/$15.52$
(354M). The consistent gains at both scales within the trained context
window confirm that value-side rotation provides a useful inductive
bias independently of any length-extrapolation effect.

\paragraph{Out-of-distribution perplexity:}
Beyond the training context, the advantage of \RoVE grows
substantially. Without positional interpolation, the 354M \RoVE
model reaches perplexity $311.38$ and $583.84$ at 4k and 16k tokens,
compared to $840.10$ and $1630.72$ for \textsc{RoPE}, i.e., a reduction of
approximately $63\%$ and $64\%$. Applying \textsc{YaRN} narrows the absolute
values for both methods, but the relative gap persists:
\RoVE+\textsc{YaRN} achieves $18.40$ and $124.82$ against $48.61$ and
$270.98$ for \textsc{RoPE}+\textsc{YaRN} at the same lengths. The 124M model
follows the same pattern (Table~\ref{tab:core_ppl_124}), with
\RoVE+\textsc{YaRN} reaching $27.34$ and $185.87$ at 4k and 16k compared
to $58.67$ and $310.52$ for \textsc{RoPE}+\textsc{YaRN}. These results
demonstrate that value-side relative rotation is complementary to, and
not subsumed by, inference-time frequency interpolation.

\paragraph{Long-context retrieval:}
Tables~\ref{tab:yarn_ruler_124} and~\ref{tab:yarn_ruler_354} report RULER
results with \textsc{YaRN} applied. At 354M, \RoVE+\textsc{YaRN} reduces the mean
RULER NLL from $6.62$ to $4.33$ relative to \textsc{RoPE}+\textsc{YaRN}, with
the largest gains on tasks requiring long-range information aggregation:
at 4k tokens, multi-key NIAH improves from $7.61$ to $3.63$ and Variable
Tracking from $4.53$ to $2.11$; at 8k tokens, the same tasks improve
from $9.46$ to $5.16$ and from $7.50$ to $3.10$, respectively. At 124M, mean
NLL decreases from $6.75$ to $5.35$, again with the strongest gains on
NIAH and Variable Tracking. The consistent pattern across both scales
and tasks indicates that positional alignment in the value pathway is
especially beneficial for retrieval problems that require detecting and
recombining information distributed across long contexts.
% --- Tables ---
\vspace{20pt}
\begin{table*}[bh]
\centering
\caption{
    Core ICL accuracy and perplexity for the 124M parameter model. Core is measured within
    the 1024-token training context; PPL is measured from 512 to 16384
    tokens. $+\textit{YaRN}$ denotes inference-time interpolation, leaving Core
    unchanged. \colorbox{bestcol}{Green highlight} marks column bests.
}
\setlength{\tabcolsep}{5pt}
\small
\begin{tabular}{lrrrrrrr}
\toprule
 &  & \multicolumn{6}{c}{PPL ($\downarrow$)} \\
\cmidrule(lr){3-8}
Method & Core ($\uparrow$) & 512 & 1024 & 2048 & 4096 & 8192 & 16384 \\
\midrule
\textsc{RoPE}         & 0.1375        & 25.23        & 22.37        & 278.97       & 500.25       & 652.28       & 808.61        \\
$+\textit{YaRN}$        &               & 25.23        & 22.37        & 22.98        & 58.67        & 175.04       & 310.52        \\
\midrule
\RoVE (ours)       & \best{0.1416} & \best{25.05} & \best{22.30} & 185.54       & 356.78       & 495.10       & 638.44        \\
$+\textit{YaRN}$       &               & \best{25.05} & \best{22.30} & \best{22.73} & \best{27.34} & \best{59.13} & \best{185.87} \\
\bottomrule
\end{tabular}
\label{tab:core_ppl_124}
\end{table*}

\begin{table*}[ht]
\centering
\caption{
    RULER long-context retrieval (NLL; lower is better) for the 124M parameter model.
    Tasks: Common Word Extraction (CWE), multi-key Needle-in-a-Haystack (NIAH),
    Question Answering (QA), Variable Tracking (VT). Avg is the unweighted mean
    over all eight task/length cells. $+\textit{YaRN}$ denotes inference-time interpolation. \colorbox{bestcol}{Green highlight} marks column bests.
}
\small
\setlength{\tabcolsep}{4pt}
\resizebox{\linewidth}{!}{%
\begin{tabular}{lccccccccc}
\toprule
 & \multicolumn{2}{c}{CWE ($\downarrow$)} & \multicolumn{2}{c}{NIAH ($\downarrow$)} & \multicolumn{2}{c}{QA ($\downarrow$)} & \multicolumn{2}{c}{VT ($\downarrow$)} &  \\
\cmidrule(lr){2-3}\cmidrule(lr){4-5}\cmidrule(lr){6-7}\cmidrule(lr){8-9}
Method & 4k & 8k & 4k & 8k & 4k & 8k & 4k & 8k & Avg ($\downarrow$)\\
\midrule
\textsc{RoPE}   & $7.56{\scriptstyle\pm0.33}$        & $8.61{\scriptstyle\pm0.40}$        & $11.57{\scriptstyle\pm1.01}$       & $11.91{\scriptstyle\pm0.96}$       & $8.81{\scriptstyle\pm1.66}$        & $8.99{\scriptstyle\pm1.56}$        & $9.72{\scriptstyle\pm0.51}$        & $10.43{\scriptstyle\pm0.57}$       & $9.70{\scriptstyle\pm0.88}$        \\
$+\textit{YaRN}$  & $4.64{\scriptstyle\pm0.49}$        & $6.13{\scriptstyle\pm0.37}$        & $7.40{\scriptstyle\pm1.70}$        & $9.79{\scriptstyle\pm1.15}$        & \best{$6.40{\scriptstyle\pm2.57}$} & $7.55{\scriptstyle\pm1.87}$        & $4.97{\scriptstyle\pm0.47}$        & $7.11{\scriptstyle\pm0.49}$        & $6.75{\scriptstyle\pm1.14}$        \\
\midrule
\RoVE (ours)   & $8.05{\scriptstyle\pm0.46}$        & $8.40{\scriptstyle\pm0.50}$        & $10.26{\scriptstyle\pm1.30}$       & $10.52{\scriptstyle\pm1.47}$       & $8.82{\scriptstyle\pm2.12}$        & $9.39{\scriptstyle\pm2.15}$        & $9.12{\scriptstyle\pm0.46}$        & $9.48{\scriptstyle\pm0.50}$        & $9.25{\scriptstyle\pm1.12}$        \\
$+\textit{YaRN}$  & \best{$4.37{\scriptstyle\pm0.53}$} & \best{$5.18{\scriptstyle\pm0.55}$} & \best{$5.09{\scriptstyle\pm1.66}$} & \best{$7.65{\scriptstyle\pm1.66}$} & $6.44{\scriptstyle\pm3.40}$        & \best{$6.80{\scriptstyle\pm2.54}$} & \best{$2.60{\scriptstyle\pm0.35}$} & \best{$4.68{\scriptstyle\pm0.46}$} & \best{$5.35{\scriptstyle\pm1.39}$} \\
\bottomrule
\end{tabular}}
\label{tab:yarn_ruler_124}
\end{table*}

\begin{table*}[ht]
\centering
\caption{
    Core ICL accuracy and perplexity for the 354M parameter model. Core is measured within
    the 1024-token training context; PPL is measured from 512 to 16384
    tokens. $+\textit{YaRN}$ denotes inference-time interpolation, leaving Core
    unchanged. \colorbox{bestcol}{Green highlight} marks column bests. \textit{Restatement of the results presented at Table~\ref{tab:core_ppl_354} for easier comparison with 124M parameter model.}
}
\setlength{\tabcolsep}{5pt}
\small
\begin{tabular}{lrrrrrrr}
\toprule
 &  & \multicolumn{6}{c}{PPL ($\downarrow$)} \\
\cmidrule(lr){3-8}
Method & Core ($\uparrow$) & 512 & 1024 & 2048 & 4096 & 8192 & 16384 \\
\midrule
\textsc{RoPE}         & 0.1664        & 17.68        & 15.64        & 293.24        & 840.10        & 1412.76       & 1630.72       \\
$+\textit{YaRN}$       &               & 17.68        & 15.64        & 16.05         & 48.61         & 154.61        & 270.98        \\
\midrule
\RoVE (ours)        & \best{0.1856} & \best{17.52} & \best{15.52} & 133.68        & 311.38        & 458.15        & 583.84        \\
$+\textit{YaRN}$      &               & \best{17.52} & \best{15.52} & \best{15.78}  & \best{18.40}  & \best{35.95}  & \best{124.82} \\
\bottomrule
\end{tabular}
\label{tab:core_ppl_354_res}
\end{table*}

\begin{table*}[ht]
\centering
\caption{
    RULER long-context retrieval (NLL; lower is better) for the 354M parameter model.
    Tasks: Common Word Extraction (CWE), multi-key Needle-in-a-Haystack (NIAH),
    Question Answering (QA), Variable Tracking (VT). Avg is the unweighted mean
    over all eight task/length cells. $+\textit{YaRN}$ denotes inference-time interpolation. \colorbox{bestcol}{Green highlight} marks column bests.  \textit{Restatement of the results presented at Table~\ref{tab:yarn_ruler_354} for easier comparison with 124M parameter model..}
}
\small
\setlength{\tabcolsep}{4pt}
\resizebox{\linewidth}{!}{%
\begin{tabular}{lccccccccc}
\toprule
 & \multicolumn{2}{c}{CWE ($\downarrow$)} & \multicolumn{2}{c}{NIAH ($\downarrow$)} & \multicolumn{2}{c}{QA ($\downarrow$)} & \multicolumn{2}{c}{VT ($\downarrow$)} &  \\
\cmidrule(lr){2-3}\cmidrule(lr){4-5}\cmidrule(lr){6-7}\cmidrule(lr){8-9}
Method & 4k & 8k & 4k & 8k & 4k & 8k & 4k & 8k & Avg ($\downarrow$)\\
\midrule
\textsc{RoPE}    & $8.87{\scriptstyle\pm0.46}$        & $9.15{\scriptstyle\pm0.44}$        & $10.90{\scriptstyle\pm1.25}$       & $10.40{\scriptstyle\pm1.22}$       & $8.83{\scriptstyle\pm1.84}$        & $9.18{\scriptstyle\pm1.64}$        & $10.12{\scriptstyle\pm0.53}$       & $10.22{\scriptstyle\pm0.38}$       & $9.71{\scriptstyle\pm0.97}$        \\
$+\textit{YaRN}$ & \best{$4.31{\scriptstyle\pm0.47}$} & $5.71{\scriptstyle\pm0.41}$        & $7.61{\scriptstyle\pm1.47}$        & $9.46{\scriptstyle\pm1.27}$        & $6.40{\scriptstyle\pm2.73}$        & $7.43{\scriptstyle\pm2.14}$        & $4.53{\scriptstyle\pm0.51}$        & $7.50{\scriptstyle\pm0.51}$        & $6.62{\scriptstyle\pm1.19}$        \\
\midrule
\RoVE (ours)    & $7.17{\scriptstyle\pm0.37}$        & $7.93{\scriptstyle\pm0.39}$        & $9.51{\scriptstyle\pm1.48}$        & $10.14{\scriptstyle\pm1.43}$       & $8.25{\scriptstyle\pm2.14}$        & $8.71{\scriptstyle\pm2.03}$        & $7.93{\scriptstyle\pm0.51}$        & $8.31{\scriptstyle\pm0.39}$        & $8.50{\scriptstyle\pm1.09}$        \\
$+\textit{YaRN}$ & $4.39{\scriptstyle\pm0.60}$        & \best{$4.62{\scriptstyle\pm0.54}$} & \best{$3.63{\scriptstyle\pm1.16}$} & \best{$5.16{\scriptstyle\pm1.21}$} & \best{$5.61{\scriptstyle\pm3.23}$} & \best{$6.05{\scriptstyle\pm2.76}$} & \best{$2.11{\scriptstyle\pm0.32}$} & \best{$3.10{\scriptstyle\pm0.41}$} & \best{$4.33{\scriptstyle\pm1.28}$} \\
\bottomrule
\end{tabular}}
\label{tab:yarn_ruler_354_res}
\end{table*}

\newpage
\clearpage
% =============================================================================
\section{Related Work}
\label{app:related}
% =============================================================================

\paragraph{Positional encodings:}
Positional encodings for transformers fall into three families.

\begin{itemize}
    \item \emph{Absolute encodings} (APE;~\citealt{vaswani2017attention}) add a fixed or
learned vector $p_i$ to each token embedding before projection, yielding scores
\begin{equation*}
  A^{\mathrm{ape}}_{ij}
  = (\Wq (x_i + p_i))\tp(\Wk (x_j + p_j)),
\end{equation*}
which depend on the absolute indices $i$ and $j$ separately, making
extrapolation to unseen lengths fragile.

\item \emph{Additive relative encodings}
(ARPE;~\citealt{raffel2020t5,press2022alibi,chi2022kerple,li2024functional})
replace the absolute positional terms with an offset-indexed bias,
\begin{equation*}
  A^{\mathrm{arpe}}_{ij} = (\Wq x_i)\tp(\Wk x_j) + b_{j-i},
\end{equation*}
so the positional contribution depends only on the displacement $\delta{=}j{-}i$, which can improve length generalisation over APE but requires materialising the full $n{\times}n$ score matrix, preventing the use of FlashAttention kernels~\citep{dao2022flashattention}.

\item \emph{Rotary position encoding}, \textsc{RoPE}~\citep{su2024roformer}, obtains the same offset-only dependence
multiplicatively: rotating queries and keys by their absolute positions before
the inner product yields
\begin{equation}
  A^{\mathrm{rope}}_{ij}
  = (\Rot{i}\Wq x_i)\tp(\Rot{j}\Wk x_j)
  = (\Wq x_i)\tp\Rot{j-i}(\Wk x_j),
  \label{eq:rope-rw}
\end{equation}
which depends on the displacement $\delta{=}j{-}i$ rather than on $i$ and $j$
separately, while remaining compatible with FlashAttention.

Length extrapolation with \textsc{RoPE} remains challenging because rotation
angles at unseen positions are out of distribution; a number of methods address
this via frequency rescaling, including PI, NTK, and \textsc{YaRN}
(see Appendix~\ref{app:freq_scl} for a brief overview).
\end{itemize}

% Length extrapolation with \textsc{RoPE} remains challenging because rotation
% angles at unseen positions are out of distribution; PI, \textsc{YaRN}, and
% LongRoPE address this via frequency
% rescaling~\citep{chen2023extendingcontextwindowlarge,peng2024yarn,ding2024longrope}.
In all three families the value pathway carries no relative-position signal: APE
injects an absolute component via $\Wv(x_j{+}p_j)$, while ARPE and \textsc{RoPE}
leave $\Wv x_j$ entirely unchanged.
\RoVE closes this gap by applying the same rotation family to the value pathway,
$\kernel{\delta}{=}\Rot{\delta}\Wv$, without modifying scores or sacrificing
FlashAttention compatibility.

\paragraph{Distinguishing ``what" and ``where" in \textsc{RoPE}:}
A separate line of work examines how position and content interact \emph{within}
the \textsc{RoPE} score itself.
\citet{barbero2024round, chen2025hope} show mechanistically that low-frequency
\textsc{RoPE} components act as semantic channels in trained models, while
high-frequency components construct positional attention patterns.
PoPE~\citep{gopalakrishnan2025decoupling} formalises this through a polar decomposition of
\eqref{eq:rope-rw}, identifying a content-dependent phase cross-term that
entangles positional and semantic information, which is replaced by pure-magnitude
representations to yield a score that factors into a content product and a
positional cosine, improving perplexity and length generalisation.
These works restructure the QK circuit to disentangle position and content in the attention mechanism; \RoVE instead introduces positional structure into the value pathway, a component none of them address.
Combining \RoVE with semantics-aware QK encodings such as PoPE is a natural
direction for future work.

\paragraph{Attention and convolution:}
\citet{cordonnier2019relationship} prove that multi-head self-attention can
express any convolutional layer under a specific relative positional encoding.
Their construction takes the ARPE score of \citet{dai2019transformer} and
zeroes the content-driven projection matrices, collapsing it to a position-only
term, i.e., a degenerate ARPE in which $b_\delta {=} v^{(h)\top} r_\delta$ absorbs
the entire score.
With the quadratic encoding $r_\delta {=} (\|\delta\|^2, \delta_1, \delta_2)$,
each head's score peaks sharply at a single fixed offset, so its value matrix
$\Wv^{(h)}$ acts as the convolutional filter for that offset.
The value pathway is thus offset-specific, but scores become content-independent
and each filter is a separate, discrete matrix tied to one head.
\RoVE reaches the same attentive-convolution structure from the opposite side:
it keeps fully content-dependent \textsc{RoPE} scores \eqref{eq:rope-rw} and
makes the value pathway offset-dependent through the continuous, parameter-free
family $\kernel{\delta} {=} \Rot{\delta}\Wv$, a single $\Wv$ rotated through
the rotation group rather than replicated per head.

DAPE~V2~\citep{zheng2025dape} takes a complementary approach, applying a
narrow convolution kernel across heads over the pre-softmax score tensor
(on top of a standard additive positional bias), and shows that this convolution
component alone provably suffices for associative recall even when the bias is
zeroed out.
However, the operation requires materialising the full $n{\times}n$ score tensor before
softmax, ruling out \textsc{FlashAttention}.
Nevertheless, the two methods are structurally dual: DAPE~V2 enriches routing while leaving
values intact, whereas \RoVE enriches the value transformation while leaving
scores intact.

\paragraph{Gated convolutions and recall:}
Gated convolutions and state-space
models~\citep{dao2022hungry,poli2023hyenahierarchylargerconvolutional,peng2023rwkv, fu2023simple} provide
sub-quadratic alternatives to attention.
\citet{arora2024zoology} show that 82\% of their perplexity gap relative to
attention is explained by \emph{associative recall}: gated convolutions apply a
fixed filter whose weights are set by model parameters alone and cannot adapt, based on the input, which tokens to mix, whereas attention's input-dependent scores
$\operatorname{softmax}(QK^\top)$ can locate the matching token at any distance.
\RoVE addresses a complementary half of this picture.
Attention already determines \emph{which} token to retrieve, but with standard
\textsc{RoPE} the value transformation $\Wv$ is the same fixed map regardless of
how far that token lies from the query.
Replacing $\Wv$ with the offset-indexed family
$\kernel{\delta} {=} \Rot{\delta}\Wv$ lets the model additionally control
\emph{how} each retrieved feature is realigned before it is recombined with the
query, much as multi-view aggregation rotates a feature into a common reference
frame before fusion \citep{miyato2024gta}.
Consistent with this view, \RoVE improves associative recall over vanilla
\textsc{RoPE} across the long-context retrieval benchmarks reported in
Appendix~\ref{app:results}.

\section{Additional Background on Frequency Scaling}
\label{app:freq_scl}

At position $t$, \textsc{RoPE} rotates frequency channel $m$ by angle
$t\,\omega_m$.
When $t$ exceeds the training context length $L$, this angle falls outside the
range $[0, L\,\omega_m]$ seen during training, causing distributional shift that
compounds across layers.
All practical remedies address this by rescaling the frequencies
$\omega_m = \theta_0^{-2m/d}$ so that positions up to a target length
$L' > L$ produce rotation angles within the training range.

\emph{Positional Interpolation} (PI;~\citealt{chen2023extendingcontextwindowlarge})
maps each position $t \mapsto tL/L'$, equivalently applying a uniform frequency
reduction $\omega_m \mapsto \omega_m / s$ with $s = L'/L$.
This guarantees all rotation angles remain in-distribution, but does so
indiscriminately: high-frequency dimensions, which encode fine-grained local
positional distinctions, are compressed by the same factor $s$ as
low-frequency dimensions, blurring short-range structure.

\emph{NTK-aware scaling}~\citep{ntk} corrects this imbalance by
uniformly rescaling the base $\theta_0$ rather than the positions directly.
Because $\omega_m = \theta_0^{-2m/d}$, a single multiplicative base change
has a dimension-dependent effect on individual frequencies: the substitution
\begin{equation*}
  \theta_0 \;\mapsto\; \theta_0 \cdot s^{d/(d-2)}
\end{equation*}
rescales each frequency as $\omega_m \mapsto \omega_m \cdot s^{-2m/(d-2)}$,
leaving the highest-frequency dimension ($m=0$) entirely unchanged while
recovering the full PI factor $s^{-1}$ at the lowest-frequency dimension
($m = d/2-1$), thereby preserving short-range positional structure where it
is most informative.

\textsc{YaRN}~\citep{peng2024yarn} takes this frequency-dependent logic to its
principled conclusion by treating each dimension according to its wavelength
$\lambda_m = 2\pi/\omega_m$ relative to the training context $L$.
Dimensions with $\lambda_m \ll L$ complete many full rotations within the
training window; their angles are robustly periodic and can therefore be safely
\emph{extrapolated} (left unscaled) at inference time.
Dimensions with $\lambda_m > L$ never complete a single rotation during
training: at positions beyond $L$ their rotation angles fall entirely outside
the training distribution, making them the primary source of extrapolation
errors. These dimensions must therefore be \emph{interpolated}.
Intermediate dimensions are handled by a smooth blend between the two regimes:
\begin{equation*}
  \omega_m' =
  \begin{cases}
    \omega_m
      & \text{if } \lambda_m < \alpha, \\[2pt]
    \bigl(1-\gamma(\lambda_m)\bigr)\,\omega_m
      + \gamma(\lambda_m)\,\omega_m/s
      & \text{if } \alpha \le \lambda_m \le \beta, \\[2pt]
    \omega_m / s
      & \text{if } \lambda_m > \beta,
  \end{cases}
\end{equation*}
where $\gamma$ is a smooth blending function increasing from $0$ to $1$
over $[\alpha,\beta]$, and $\alpha,\beta$ are wavelength thresholds.
To compensate for the softmax sharpness distortion introduced by frequency
compression, \textsc{YaRN} additionally applies an attention temperature
correction $\sqrt{1/t}$ to the pre-softmax logits.
All adjustments are applied at inference time without any additional fine-tuning.

More advanced frequency extrapolation schemes have since been
proposed~\citep{ding2024longrope, tian2026mrropemixedradixrotaryposition}; however, in this work we focus on
\textsc{YaRN} as it is among the most widely adopted techniques in production
settings, as evidenced, for instance, by its use in
DeepSeek-V4~\citep{deepseekai2026deepseekv4}.

\paragraph{Frequency scaling in \RoVE:}
When applying \textsc{YaRN} to \RoVE, we rescale the frequencies of both the
QK and OV rotation matrices uniformly, as described above.
This is a natural extension: since \RoVE ties the value pathway to the same
rotation family $\{\Rot{\delta}\}_\delta$ as the keys and queries, the same
out-of-distribution problem arises in the OV circuit when positions exceed $L$,
and the same frequency rescaling mitigates it.
We leave as future work whether more specialised extrapolation strategies
should be developed specifically for the OV pathway. The methods reviewed
above, such as, PI, NTK-aware scaling, and \textsc{YaRN}, were originally designed
under the assumption that rotations appear only in the QK circuit; it is
therefore an open question whether their frequency thresholds and blending
schedules remain optimal when the same rotation family also modulates value
transformations, or whether the two pathways call for different rescaling
regimes.
% % =============================================================================
% \section{Implementation Details}
% \label{app:implementation}
% % =============================================================================

% \RoVE reuses existing \textsc{RoPE} kernels without modification.
% After projecting to values $v_j = \Wv x_j \in \R^d$,
% reshaping to $\R^{\frac{d}{2}\times2} \cong \mathbb{C}^{\frac{d}{2}}$ and
% multiplying elementwise by $e^{ij\omega_k}$ (the standard \textsc{RoPE} rotation);
% the post-aggregation inverse at position $i$ multiplies by $e^{-ii\omega_k}$.
% The combined effect is $e^{i(j-i)\omega_k}$, implementing $\Rot{j-i}$.
% For FlashAttention compatibility: rotate $V$ before calling the
% attention kernel, rotate output $O$ by the query's inverse rotation after.
% Both operations are $\mathcal{O}(nd)$ and require no kernel modification.

% % =============================================================================
\section{Discussion}
\label{sec:discussion}
% =============================================================================
\paragraph{What changes when values rotate?}
In standard \textsc{RoPE}, relative position determines attention weights but
leaves the value map $\Wv$ unchanged: position governs \emph{which} features
are aggregated, not \emph{how}. \RoVE closes this gap by replacing $\Wv$ with
the offset-dependent kernel $\Rot{\delta}\Wv$, so that the layer remains
shift-equivariant while the value stream carries strictly more
relative-position information than a scalar attention coefficient can convey
alone. From a geometric perspective, attention selects which neighboring
features to aggregate while $\Rot{\delta}\Wv$ aligns each selected feature
before fusion, analogous to geometric multi-view aggregation
in~\cite{miyato2024gta}, where features from different camera views are rotated
into a common frame; in language, positions replace views and relative-position
rotations replace camera-to-camera transforms. This joint conditioning of
selection and aggregation explains the RULER pattern, where gains are largest
on tasks requiring long-range information to be tracked and recombined rather
than merely detected.

\paragraph{Limitations and future work:}
While the experiments show a consistent advantage for rotating values, they do
not fully identify the mechanism behind the improved OOD perplexity. Our
working hypothesis is that \RoVE induces a more coherent extrapolation
regime: when relative offsets exceed those seen in training, the QK and value
pathways drift together because they share the same rotation family. In
standard \textsc{RoPE}, by contrast, the attention logits extrapolate while
the value transformation remains unchanged, creating a mismatch between
selection and aggregation. A direct test would analyze errors by \textsc{RoPE}
frequency band, examining each term in the offset-indexed sum of
Appendix~\ref{app:circuits} separately, following the style
of~\cite{chen2025hope}, and measure whether value-side rotations
preserve the learned relationship between the QK and OV circuits at unseen
offsets.

% =============================================================================
\section{Circuits Framework Analysis}
\label{app:circuits}
% =============================================================================

\paragraph{Background:}
\cite{elhage2021mathematical} provide a notation for decomposing transformer
computations into interpretable end-to-end paths. We recall the elements
needed here.

The \emph{token embedding matrix} $W_E \in \R^{d \times |\mathcal{V}|}$ maps
one-hot token vectors to residual-stream vectors of dimension $d$; the
\emph{unembedding matrix} $W_U \in \R^{|\mathcal{V}| \times d}$ maps
residual-stream vectors back to logits over the vocabulary $\mathcal{V}$.
Together they form the ``direct path'' from input tokens to output logits:
$\mathrm{Id} \otimes W_U W_E$, where $\mathrm{Id}$ is the identity on the
sequence dimension and $\otimes$ denotes the tensor product (equivalently, a
Kronecker product when written on vectorised tokens).

\begin{figure}[t]
\floatconts
  {fig:subfigex}
  {\vspace{-15pt}\caption{Visualization of the circuits of the one-layer attention-only transformer for \textsc{RoPE} (left; adapted from \cite{elhage2021mathematical}) and \RoVE (right). In standard \textsc{RoPE}, we have as many bifurcations from the residual stream as there are attention heads $h_i$ at that layer. In \RoVE we obtain new bifurcations associated with each displacement $\delta$.}}
  {%
    \setlength{\jmlrminsubcaptionwidth}{0.26\linewidth}%
    \subfigure[\centering \textsc{RoPE} attention circuit see Equation~\eqref{eq:circuits-rope}]{\label{fig:circuits-rope}%
      \includegraphics[width=0.28\linewidth]{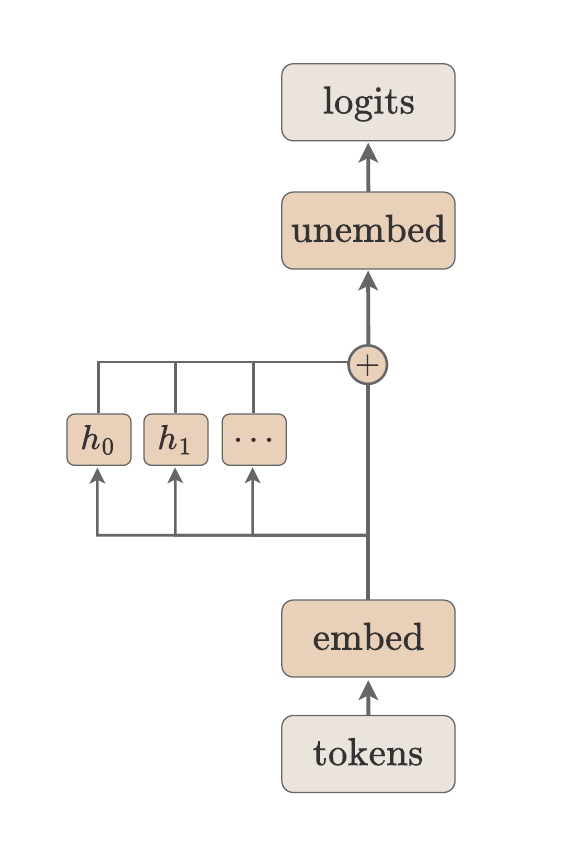}}%
    \qquad\qquad
\subfigure[\RoVE attention circuit; see Equation~\eqref{eq:circuits-RoVE}]{\label{fig:circuits-Rove}%
  \includegraphics[width=0.55\linewidth]{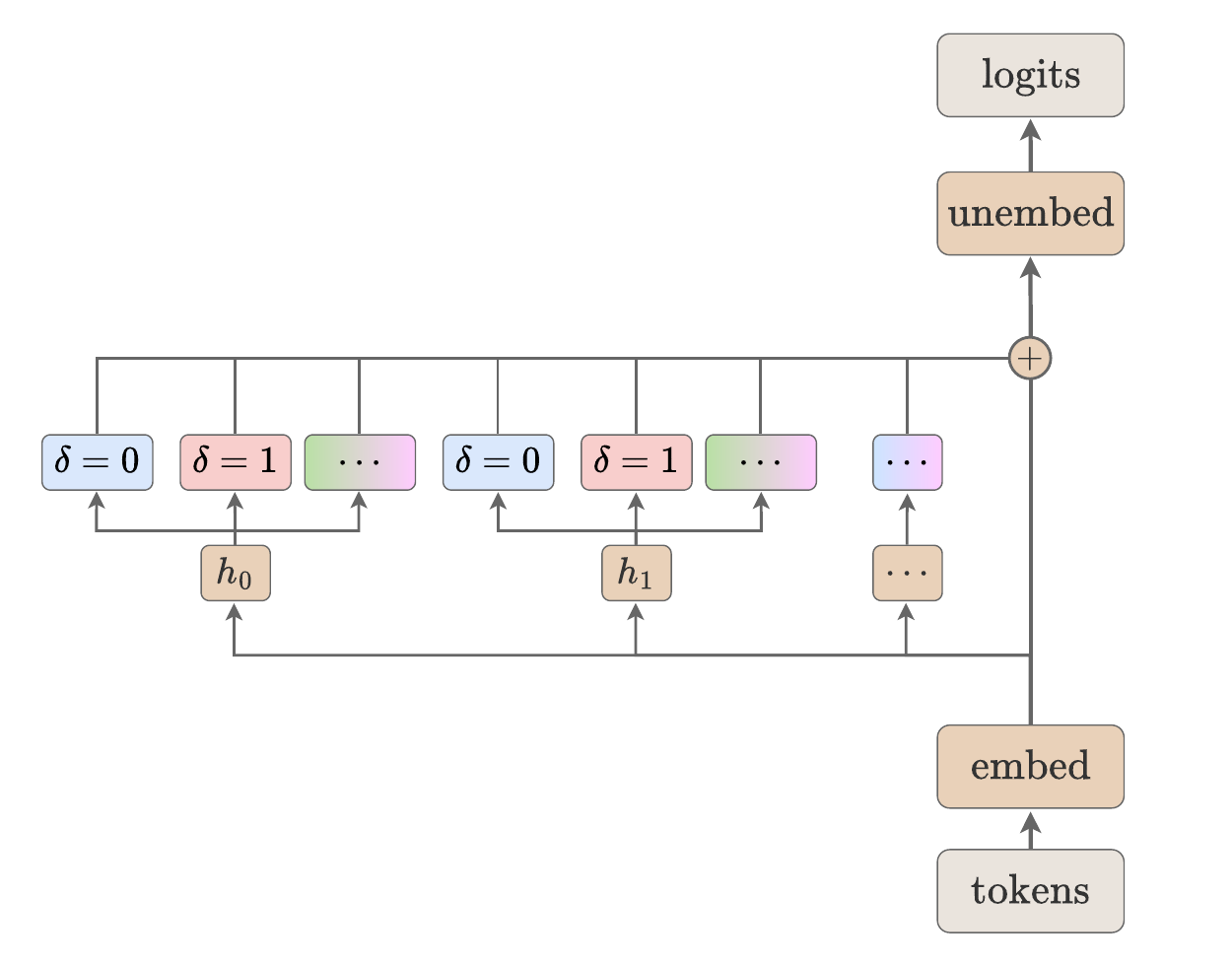}}%}%}%
  }
\end{figure}

Each attention head $h$ contributes through two largely independent circuits:
\begin{itemize}
\item \textbf{QK circuit.} The bilinear form
$W_{QK}^h = (\Wq^h)\tp\Wk^h \in \R^{d\times d}$
determines the attention pattern $A^h(X) \in \R^{n\times n}$: entry
$A^h(X)_{ij}$ measures how strongly position $i$ attends to position $j$
based on the residual-stream content $X$ at those positions. It answers the
question \emph{which} tokens are attended to.

\item \textbf{OV circuit.} The matrix
$W_{OV}^h = \Wo^h\Wv^h \in \R^{d\times d}$
determines what is communicated when a token is attended to: it maps the
residual-stream vector at position $j$ to the update written into position
$i$'s residual stream. It answers the question \emph{what} information is
moved.
\end{itemize}
The full one-layer attention-only transformer then expands as
\begin{equation}
  T(X) = \mathrm{Id}\otimes W_U W_E
  \;+\;
  \sum_{h} A^h(X) \otimes \bigl(W_U W_{OV}^h W_E\bigr),
  \label{eq:circuits-rope}
\end{equation}
where the tensor product $A^h(X) \otimes (W_U W_{OV}^h W_E)$ means: $A^h(X)$
routes information across sequence positions in an input-dependent way, while
$W_U W_{OV}^h W_E$ transforms it in the channel dimension with fixed weights.
The two dimensions are \emph{independent}---this is the Kronecker structure. See Figure~\ref{fig:circuits-rope} for a visual representation of Equation~\eqref{eq:circuits-rope}.

\paragraph{How RoPE fits in:}
With \textsc{RoPE}, the attention pattern $A^h(X)$ becomes shift-equivariant
(entry $A^h(X)_{ij}$ depends on positions only through the offset $j-i$ and
the token content), but the OV circuit $W_{OV}^h$ remains a single fixed
matrix. The Kronecker structure of~\eqref{eq:circuits-rope} is therefore
preserved under \textsc{RoPE}: token routing and channel transformation are
still decoupled.

\paragraph{How \RoVE changes the picture:}
\RoVE operates at the attention level, replacing the constant value
map $\Wv^h$ with the offset-dependent kernel $\Rot{\delta}\Wv^h$. From the
circuits perspective, this means the effective OV matrix at offset $\delta$ is
$\Wo^h\Rot{\delta}\Wv^h$, with the rotation $\Rot{\delta}$ sandwiched between
the two projections.
Consequently, the token-routing and channel-transformation dimensions are no
longer independent, and~\eqref{eq:circuits-rope} no longer holds. Introducing
shift matrices $\shift{\delta}\in\R^{n\times n}$ with
$(\shift{\delta})_{ij}=\mathbf{1}[j-i=\delta]$ to partition the attention
pattern by offset, the \RoVE transformer expands as
\begin{equation}
  T^{\RoVE}(X) = \mathrm{Id}\otimes W_U W_E
  \;+\;
  \sum_{h}\sum_{\delta}
  \bigl(A^h(X)\odot\shift{\delta}\bigr)
  \otimes
  \bigl(W_U \Wo^h \Rot{\delta} \Wv^h W_E\bigr),
  \label{eq:circuits-RoVE}
\end{equation}
where $\odot$ is the entry-wise product. Each term selects the $\delta$-offset
entries of $A^h(X)$ and pairs them with the corresponding rotated end-to-end
map, and summing over $\delta$ recovers the full output because
$\sum_\delta A^h(X) \odot \shift{\delta} = A^h(X)$. The Kronecker structure
is replaced by a \emph{sum of Kronecker products}, one per offset diagonal,
which is exactly the gated block-Toeplitz structure described in Section~\ref{sec:RoVE}:
the gate $A^h(X)\odot\shift{\delta}$ selects each offset diagonal of an
otherwise block-Toeplitz kernel. See Figure~\ref{fig:circuits-Rove} for a visual representation of Equation~\eqref{eq:circuits-RoVE}.

\end{document}